\documentclass[graybox]{svmult}

\usepackage{type1cm}        
\usepackage{makeidx}         
\usepackage{graphicx}        
\usepackage{multicol}        
\usepackage[bottom]{footmisc}

\usepackage{cite}
\usepackage{hyperref}
\usepackage{wrapfig}

\usepackage{newtxtext}       %
\usepackage[varvw]{newtxmath}       

\makeindex             

\begin{document}

\title*{Towards Conscious Service Robots}
\author{Sven Behnke\orcidID{0000-0002-5040-7525}}
\institute{Sven Behnke \at Autonomous Intelligent Systems, Computer Science
Institute VI -- Intelligent Systems and Robotics, Center for Robotics
and the Lamarr Institute for Machine Learning and Artificial Intelligence, University of Bonn, Germany. \email{behnke @ cs.uni-bonn.de}
}
%
%
\maketitle

\abstract*{Deep learning's success in perception, natural language processing, etc. inspires hopes for advancements in autonomous robotics. However, real-world robotics face challenges like variability, high-dimensional state spaces, non-linear dependencies, and partial observability. A key issue is non-stationarity of robots, environments, and tasks, leading to performance drops with out-of-distribution data. Unlike current machine learning models, humans adapt quickly to changes and new tasks due to a cognitive architecture that enables systematic generalization and meta-cognition. Human brain's System~1 handles routine tasks unconsciously, while System~2 manages complex tasks consciously, facilitating flexible problem-solving and self-monitoring. For robots to achieve human-like learning and reasoning, they need to integrate causal models, working memory, planning, and metacognitive processing. By incorporating human cognition insights, the next generation of service robots will handle novel situations and monitor themselves to avoid risks and mitigate errors.}

\abstract{Deep learning's success in perception, natural language processing, etc. inspires hopes for advancements in autonomous robotics. However, real-world robotics face challenges like variability, high-dimensional state spaces, non-linear dependencies, and partial observability. A key issue is non-stationarity of robots, environments, and tasks, leading to performance drops with out-of-distribution data. Unlike current machine learning models, humans adapt quickly to changes and new tasks due to a cognitive architecture that enables systematic generalization and meta-cognition. Human brain's System~1 handles routine tasks unconsciously, while System~2 manages complex tasks consciously, facilitating flexible problem-solving and self-monitoring. For robots to achieve human-like learning and reasoning, they need to integrate causal models, working memory, planning, and metacognitive processing. By incorporating human cognition insights, the next generation of service robots will handle novel situations and monitor themselves to avoid risks and mitigate errors.}

\section{Introduction}
\label{sec:Intro}

Industries like car manufacturing impressively demonstrate the utility of robots. Recent developments in sensing, actuation, and -- most importantly -- artificial intelligence (AI) make it conceivable that robots will revolutionize many new application domains such as the flexible production of small lots, logistics, agriculture, security, inspection, professional services, and personal assistance. All of these domains, however, require, cognitive capabilities far beyond those of today’s robots. 

Current robotic systems rely on structuring the environments and tasks, e.g. by providing objects in well-defined locations. In open-ended settings, such as our everyday environments, robot-friendly structuring is impossible. Instead, autonomous service robots must instantiate models of their environment from sensor measurements, plan actions to achieve goals, carry out plans in the presence of disturbances, and monitor their execution. They also must familiarize with new objects and tools and need to improve their behavior through learning. Finally, they must communicate with persons in a human-understandable way, to receive instructions, answer questions, and explain their behavior.

The tremendous success of deep learning~\cite{SchulzB12,LeCun15} in visual perception, speech recognition, natural language processing, vision-language tasks, and multimodal tasks gives rise to hope that these methods will lead to revolutionary advances in autonomous robot performance. 



\section{Related Work}
\label{sec:Related}

\runinhead{Personal service robots} Personal robots that assist handicapped or elderly persons in their activities of daily living have attracted much attention in robotics research. An increasing number of research groups are working on robots for service applications. Examples include PR2~\cite{MeeussenWGCMMMMEFHRMBKGB10} that was used in a household marathon experiment~\cite{KazhoyanSKKB21}, Everyday Robots’ mobile manipulator~\cite{HerzogRHLWYLAXK23}, and Toyota Human Support Robot~\cite{YamamotoTOIIAYI19}, a standard platform in the international RoboCup@Home competitions. My team NimbRo won these competitions 2024 with two PAL Robotics TIAGo++ robots~\cite{Memmesheimer25} and 2011--2013 with our cognitive service robot Cosero~\cite{StuecklerRAM12,StucklerSB16}, demonstrating a large variety of domestic service tasks. Further examples include Care-O-Bot 4~\cite{KittmannFSRWH15}, Armar-6/7~\cite{AsfourPWKRWOGZG19}, HRP-5P~\cite{KumagaiKMSNKKKB19}, TORO~\cite{KheddarRWCSOLEC19}, E2-DR~\cite{YoshiikeKKUKKHK19}, and our Centauro robot~\cite{KlamtKKKLLLHMPP19,KlamtSLBBCDDGKK20} which demonstrated challenging locomotion and manipulation tasks including the use of tools. For the ANA Avatar XPRIZE competition~\cite{BehnkeAL23, Hauser24}, capable systems have been developed, including iCub3~\cite{DafarraPRRGDMVSVSTSEGHLDMMP24}, Pollen Robotics Reachy, and our winning Avatar system~\cite{SchwarzLMPRSB23,LenzSchwarz:SORO2023}.
Scene perception. In order to act in complex indoor environments, service robots must perceive the room structure, obstacles, persons, objects, etc. To this end, they are equipped with cameras and depth sensors. Estimating the sensor poses and registering the measurements yields environment maps~\cite{PlacedSCAICC23}. In addition to modeling the environment geometry and appearance, semantic perception is needed~\cite{Schmid24}.

\runinhead{Deep learning} For pattern recognition, deep learning~\cite{LeCun15} methods are extremely successful. They revolutionized visual perception~\cite{Dehghani23, Ravi24}, speech recognition~\cite{RadfordKXBMS23, PratapTSTBKENVF24}, natural language processing~\cite{OpenAI23, Anil23}, vision-language tasks~\cite{LiuLWL23a, Sun24}, and multimodal tasks~\cite{GirdharELSAJM23, HanGZ0ZL24}. Supervised deep learning requires large annotated data sets like ImageNet-21K~\cite{DengDSLL009}, JFT-3B~\cite{Zhai0HB22}, and Kinetics~\cite{CarreiraZ17}, though, which are expensive to obtain. To address variability that should not change output, data augmentation methods such as image transformations~\cite{ShortenK19}	and generative models~\cite{YuXTSWBSTMPHIX23} are used to generate variants of training examples. For robotic tasks such as mobile manipulation, large-scale annotated datasets do not exist. To avoid the need for large labeled datasets, much research focuses on methods that can adapt to new conditions through transfer learning and domain adaptation. Transfer learning~\cite{KolesnikovBZPYG20} uses representations learned from large data to learn a related task from small data, e.g. by continuation of training. Semi-supervised, weakly supervised, and unsupervised learning methods use fewer, low quality, and no labels at all, respectively. One example of semi-supervised methods is the student-teacher approach~\cite{XieLHL20}, where a teacher is trained on a small labeled data set and then generates pseudo labels for a large unlabeled data set to train the student. Because unlabeled data is much easier to obtain than annotated data, unsupervised methods are often used to pre-train models~\cite{BengioCV13, ChenY0OL23}. The hope is to discover useful structure in the data which might aid target tasks. A promising subclass of unsupervised learning is self-supervised learning~\cite{GuiCZCSLT24}, which requires only unlabeled data to formulate a pretext task, for which a target objective can be computed without supervision. These pretext tasks must be designed in a way that high-level data understanding is useful for solving them, e.g. prediction of occluded image parts~\cite{HeCXLDG22} or future video frames. As a result, the intermediate layers of trained models encode high-level semantic representations that are useful for solving downstream tasks. One form of self-supervision is contrastive learning~\cite{ChenK0H20}, where two different data augmentations are applied to an image and a model is trained to maximize agreement between the outputs and minimize agreement with outputs for other images. Contrastive learning of dense descriptors for object surface elements has been applied to learn visuomotor manipulation policies~\cite{FlorenceMT20} that generalize within a category of objects and are able to handle deformable objects. Other possibilities for self-supervised learning are to maximize mutual information between input and model output~\cite{HjelmFLGBTB19} and joint embeddings of two inputs with variance-invariance-covariance regularization~\cite{AssranDMBVRLB23}. 

\runinhead{Large language models} Self-supervised training is the basis for the impressive performance of recent large language models (LLMs) such as GPT-4~\cite{OpenAI23} and Palm~2~\cite{Anil23} that continue text in plausible ways. Their large transformer networks~\cite{VaswaniSPUJGKP17} were trained on massive data to predict the next token. In contrast to recurrent sequence models, transformers flexibly re-route and combine information from relevant parts of the sequence through learned self-attention, which is implemented using content-based access to information values by matching keys to queries. Recently, autoregressively trained LLMs have shown sparks of artificial general intelligence~\cite{Bubeck23}. Such models can acquire human-like systematic generalization through meta-learning~\cite{LakeB23}, but this requires generating a training set of systematic generalization example problems. On the other hand, LLMs often lack common sense, hallucinate facts, fail at arithmetic, have difficulty reasoning, and cannot make proper plans. For these reasons, LLMs are combined with external tools~\cite{QinLYZYLLCTQZHT24} such as search engines, calculators, planners~\cite{Liu23}, etc. Multimodal models such as PaLM-E~\cite{DriessXSLCIWTVY23} and ImageBind~\cite{GirdharELSAJM23} combine text and other modalities such as images in joint embeddings. Generative models are not restricted to producing text but are also used to generate, e.g., images~\cite{PerniasRRPA24} and video~\cite{Qin24} from text inputs.

\runinhead{3D models} To address the 3D nature of scenes, Neural Radiance Fields (NeRF) have been proposed, which learn a neural network mapping 5D coordinates to density and color by predicting images from multiple views through volumetric rendering~\cite{MildenhallSTBRN22}. In our recent work PermutoSDF~\cite{RosuB23}, the 3D shape of objects is represented by a neural signed distance function (SDF). By modeling individual objects as permutation-invariant slots, object representations can be learned through novel-view synthesis~\cite{SajjadiDMSPLGGK22}. If conditioned on latent variables, category-level shape spaces can be learned, e.g., for articulated human bodies~\cite{DengLJPH0T20}. Compositional generative scene models that represent objects and their relations can be learned without image-level supervision~\cite{GaoLC0S24}.

\runinhead{Scene prediction} In dynamic scenes, the motion of objects and persons must be estimated and predicted. Scenes with moving agents (e.g., humans or robots) can be represented with 3D dynamic scene graphs~\cite{HughesCC22}. Motion is the strongest cue for perceptual grouping and predictive models are widely used to explain human visual perception~\cite{Friston18}. Consequently, optimization of a prediction loss can be used to segment moving objects in videos~\cite{Wang24}. SlotFormer~\cite{WuDGKG23} models spatio-temporal object relationships and predicts object states. Our recent work on object-centric video prediction decouples the processing of temporal dynamics and object interactions~\cite{VillarCorralesWB23}. This facilitates learning of tasks that require understanding of object relations~\cite{MosbachSOLD24}. A fundamental problem when predicting the future is that often multiple plausible futures exist. MultiPath++~\cite{VaradarajanHSRN22} predicts a distribution of future trajectories of road users parameterized as a Gaussian Mixture Model (GMM). Multiverse~\cite{Liang20} predicts the distribution over multiple possible future paths of persons using convolutional recurrent neural networks (RNNs) over graphs. 
World modeling. Prediction of future scene states and planning own actions require world models that are conditioned on actions. Playable Video Generation~\cite{MenapaceLTS021} learns a discrete set of actions from unlabeled video that are used to interactively generate video from actions. This task has been extended to Playable Environments~\cite{MenapaceLTS021} that can control multiple objects in 3D scenes with action labels that are discovered in an unsupervised way. DayDreamer~\cite{WuEHAG22} learns action-conditioned forward models in a latent space for multiple robots. GenAD~\cite{YangGQ0L0CWZ0Z024} and GameNGen~\cite{Valevski24} are world models for autonomous driving and a video game, respectively.

\runinhead{Deep reinforcement learning} Reinforcement learning (RL) addresses the development of situated agents that learn how to behave while interacting with the environment~\cite{SuttonBarto2018}. This problem is formulated as an agent-centric optimization in which the objective is to select actions based on the estimated state in order to obtain as much reward from the environment as possible in the long run. Impressive success has been achieved by combining this approach with deep learning. One example is MuZero~\cite{SchrittwieserAH20} which combines tree-based search with a learned model and achieves superhuman performance in a range of challenging and visually complex domains (Atari games, Go, chess, and shogi), without any prior knowledge of their underlying dynamics. MuZero learns a model that predicts the quantities relevant to action planning: the reward, the action-selection policy, and the value function. AlphaStar~\cite{VinyalsBCMDCCPE19} learned the multi-agent game StarCraft II from $\sim$500K human games and 120M self-played games. Gran Turismo Sophy~\cite{WurmanBKMS0CDE022} learned from carefully engineered state and reward in more than five years of simulated driving hours to compete with the world’s best drivers. Playing soccer with humanoid agents was learned from decades of match simulations~\cite{LiuLWMEHCTOASHM22}. Soccer skills for a humanoid robot and 1v1 play were learned from 2.5\,years of simulated experience~\cite{HaarnojaMLHTHWTSHBHBHTSBCSG24}.

 These numbers indicate that it would be impractical to collect that much experience with a real robotic system. Consequently, real-robot reinforcement learning mostly focuses on individual skills. For example, Google X learned grasping from cluttered bins with a simple manipulator under closed-loop monocular vision-based control~\cite{KalashnikovIPIH18}. They operated seven experimental setups for four months to collect 580K real-world grasp attempts to train a deep neural network Q-function with over 1.2M parameters and report a 96\% grasp success rate on unseen objects. The method learned regrasping strategies, probing or repositioning objects to find the most effective grasps, performing other non-prehensile pre-grasp manipulations, and responding dynamically to disturbances and perturbations. Sorting recyclables and trash was learned from simulation and 9,527\,hours of real-robot experience obtained with a fleet of 23 mobile manipulators~\cite{HerzogRHLWYLAXK23}.

 Real-robot RL needs suitable inductive biases~\cite{Hessel19} to learn from little experience. These biases represent domain knowledge and can take many forms, e.g., the structure of the agent-environment interface and the policy generation mechanism. To improve the data efficiency of RL, transfer learning has been investigated. By pre-training on RoboNet~\cite{DasariETNBSSLF19}, a data set providing 15M video frames from seven different robot platforms, and fine-tuning on a held-out target platform, it has been demonstrated that simple manipulation tasks such as pushing and pick-and-place can be learned from limited experience. Multi-task learning amortizes experience over multiple tasks~\cite{KalashnikovVCSJ21}. It generalizes to structurally similar tasks and acquires distinct new tasks more quickly. 
One way to address the combinatorial complexity of multi-object scenes is to factorize them into objects. Object-centric perception, prediction, and planning~\cite{VeerapaneniC0JF19} learns to discover objects in visual scenes and models their dynamics and appearance without supervision. A model-based reinforcement learner that predicts and plans block stacking on this abstract level generalizes to novel configurations and more objects. Action Schema Networks~\cite{ToyerTTX18} learn generalized policies for probabilistic planning problems. By mimicking the relational structure of planning problems, they generalize over all instances of a given planning domain. Manipulation inherently involves contact and often requires both haptic and visual feedback. Lee et al.~\cite{LeeZZTSSFGB20} use self-supervision to learn a compact and multimodal representation of sensory inputs, which is then used to improve the sample efficiency of policy learning as demonstrated for peg insertion. 
To avoid random exploration, imitation of human experts can be used. RT-1~\cite{BrohanBCCDFGHHH23} is a transformer-based controller trained on 130K demonstrations of a large variety of pick and place tasks in kitchen environments. Open\,X-Embodiment~\cite{ONeillRMGPLPGMJ24} is a large data set of camera images and end-effector movements from 22 different robots, demonstrating 527 skills (160,266 tasks). The RT-X model trained on this data exhibits positive transfer and improves the capabilities of multiple robots by leveraging experience from other platforms. 970k episodes from this data set were used to train OpenVLA~\cite{Kim24}, starting from a large language model and a visual encoder. OpenVLA demonstrates generalist manipulation capabilities and can be adapted to new robots via fine-tuning.

\section{Challenges}
\label{sec:Challenges}

Despite much research and progress, capable mobile manipulation robots that can cope with the complexity of open-ended real-world applications have not yet been realized. Developing such robots is a tremendous challenge, due to the typical characteristics of these applications. 

\runinhead{Many sources of variability} There are many sources of variability that a mobile manipulation robot must cope with. These include varying shape, texture, and physical properties of objects -- even within a category. Furthermore, the 6D object pose, speed, and articulation state may vary. Environmental conditions, such as lighting, and surface properties, such as shininess, transparency, texturelessness, or non-reflectivity greatly impact appearance in camera images and consequently the completeness and precision of depth estimates. The variability of single objects is exponentiated by the infinite possibilities for multi-object arrangements. Similarly, the robot environments such as rooms and apartments vary greatly in layout, geometry, surface properties, and other factors. The manipulation and locomotion tasks that capable robots need to perform are highly variable as well. Hence, learning methods are needed that generalize to novel, unseen situations.     

\runinhead{High-dimensional state and action spaces} Input and output of mobile manipulation robot controllers are high-dimensional. Typical camera images are of size 1920$\times$1080$\times$3, already more than 6M dimensions. Depth cameras, 3D LiDARs, force-torque \& haptic, inertial, and joint sensors add many more input dimensions. The sensors measure at high rates, e.g., at 30\,Hz, producing hundreds of million measurements per second. The output dimensionality is high as well, with typically more than 50\,DoF for anthropomorphic robots. These joints need to be controlled at high rates with target positions, velocities, or torques. Hence, learning methods are needed that can cope with high-dimensional state and action spaces.

\runinhead{Hybrid discrete-continuous variables} Some variables, such as the presence of objects or the task category, are discrete while other variables, such as 6D object poses or task parameters, are continuous. This creates the need for learning methods that can cope with both discrete and continuous variables.

\runinhead{Non-linear dependencies} Objects are typically in contact with support surfaces and with each other and the robot must make and release contacts with its end-effectors or other body parts to manipulate them. This induces highly non-linear constraints. While objects may be easily moved away from the contact point, moving them further towards the colliding surface is not possible. Similarly, occlusion effects and the transition between stick friction and sliding are highly non-linear. Learning methods must address such non-linearities.

\runinhead{Stochasticity} There is much randomness in the world. Unmodeled environmental factors and other agents might also be perceived as non-determinism from our robot’s point of view. Furthermore, robot sensors are noisy and unreliable; and robot actuators are imperfect and induce stochasticity. Hence, the state must be estimated from unreliable observations and predictions are hard to make and become more and more uncertain for larger time horizons. Learning methods must cope with such uncertainties.  

\runinhead{Partial observability} Due to the projection of the 3D world onto 2D cameras and other sensors, limited sensor ranges, resolutions, accuracies, etc., not all state variables that would be needed for action planning are directly accessible. Hence, learning methods must consider the distribution of possible states and must generate actions to acquire more information, for example changing the camera pose to see occluded objects, touching objects to sense physical properties like weight and stiffness, and opening containers to see what is inside. 

\runinhead{Underactuation} Robots have limited action capabilities to influence the state of the environment. Their drives have limited speed and acceleration, their manipulators have limited reach, strength, and dexterity. Some environmental variables cannot be influenced directly, but only through indirect means like tools. Learning methods must respect these constraints and generate behavior such as improvised tool use to overcome them. 

\runinhead{Multimodality} Mobile manipulation involves multiple modalities, such as vision, distance measurements, forces, and haptics on the input, and also multiple outputs, such as mobility, manipulation, and active sensing. Hence, learning methods must jointly address these modalities and come up, for example, with grasping strategies that transition smoothly from vision-based scene understanding and grasp selection, to visual tracking and correction of the approaching motion, to grasping execution and re-grasping based on haptic feedback. 
 
\runinhead{Non-stationarity} One unique challenge is the non-stationarity of robots and their environments. Not only do robot bodies change due to wear and tear, also the open-ended environments in which they operate and the tasks they perform are constantly changing. Already the ancient philosopher Heraclitus noted that the only constant in life is change. Such changes violate the fundamental assumption underlying current machine learning that a learned model will be used on the same distribution of data it has been trained on. When using a trained model on a different distribution (out-of-distribution, OOD), one cannot expect good performance~\cite{MorenoTorresRACH12}. In fact, seemingly small changes can lead to catastrophic failure~\cite{ChenLWJDZ23}.

\section{Human Cognitive Functions}

Humans are able to cope with such changes and quickly learn new tasks. My hypothesis is that the cognitive architecture of the human mind has evolved to continuously interact with changing environments and that equipping robots with key elements of this architecture will enable flexible handling of OOD data and \textit{systematic generalization}. Systematic generalization was first studied in linguistics~\cite{LakeB:ICML18, BahdanauMNNVC:ICLR19} because it is a core property of language that meaning for a novel composition of existing concepts (e.g. words) can be derived systematically from the meaning of the composed concepts and the way they are composed. Humans exhibit systematic generalization also when understanding a new object by combining properties or parts which compose it~\cite{Lake:BBS2017}. \textit{Compositionality} is the principle that complex objects can be described by their constituent parts and their relations to each other~\cite{Fodor:2001}. It allows to generate infinite variants from a finite set of building blocks, enables open-world zero-shot learning~\cite{ManciniNXA:CVPR21}, and even makes it possible to generalize to new combinations that have zero probability under the training distribution. 

\begin{wrapfigure}{r}{0.65\textwidth}
    \centering\vspace*{-3ex}
         \includegraphics[width=7.5cm]{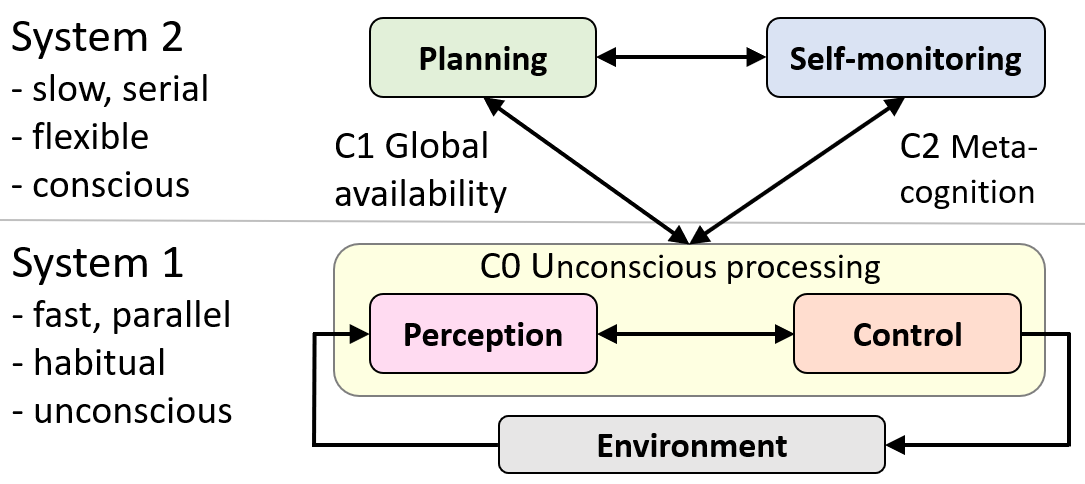}\vspace*{-1ex}
        \caption{Human cognitive functions according to Kahneman~\cite{kahneman2011thinking} (System~1, System~2) and Dehaene et al.~\cite{Dehaene:2017} (C0,  C1, C2).}
				\vspace*{-2ex}
        \label{fig:CogArch}
\end{wrapfigure}

While humans perform many routine tasks like walking or riding a bike without much attention, object manipulation, communication, and handling novelty are different. Cognitive science distinguishes \textit{habitual} and \textit{controlled} processing~\cite{Botvinick:2001}. Habitual processing effortlessly generates default behaviors that are performed routinely. In contrast, controlled processing requires attention and mental effort to generate non-routine behaviors. Kahneman~\cite{kahneman2011thinking} introduced the framework of fast and slow thinking and corresponding processing systems in our brain (see Fig.~\ref{fig:CogArch}). Routine, habitual tasks can be achieved quickly in parallel without \textit{conscious} attention using only System~1 abilities, whereas more complex tasks also require System~2 that is more capable but slower, serial and involves conscious processing. System~2 uses explicit, verbalizable knowledge and explicit processing while System~1 relies on implicit, non-verbalizable, intuitive knowledge. We can act in fast and precise habitual ways without having to think consciously, but the reverse is not true: conscious processing builds on the unconscious System~1. 

System~2 is very flexible and powerful. It allows to solve novel problems creatively by recombining existing pieces of knowledge, to discover and use causal dependencies, to imagine future outcomes, to plan actions, to find explanations, to reason, etc. It is also at that level that we communicate with others through natural language, e.g. to receive task specifications or new knowledge and rules that we can apply immediately. The capacity of System~2 is very limited, though. Our working memory can only hold 3-5 meaningful items active simultaneously~\cite{Cowan:2010}. Baars introduced the \textit{Global Workspace Theory}~\cite{Baars:1993} that identifies conscious processing as the communication bottleneck between selected parts of the brain that are called upon when addressing a current task. There is a threshold of relevance beyond which information that was previously handled unconsciously gains access to this bottleneck and is instantiated in working memory. When this happens, the information is broadcast throughout the brain, allowing its various relevant parts to synchronize and choose configurations and interpretations of their piece of information that are globally coherent with the configurations chosen in other parts of the brain. 

While this severe communication bottleneck might appear to be a weakness, it can also be advantageous. Firstly, there is pressure to combine multiple lower-level items that frequently occur together to larger, composite items, facilitating abstracting away irrelevant detail and providing compositionality. Secondly, when focusing on a few relevant items of a scene for conscious planning, we essentially ignore all other items, which are irrelevant for the task at hand. This leads to systematic generalization, because we can reuse the task knowledge in infinitely many novel situations in which the irrelevant items change. System~2 is slow, has limited capacity, and involves conscious effort; hence, there is pressure to migrate tasks to System~1 wherever possible. Through rehearsal, frequently performed tasks become habitual. 

Dehaene et al.~\cite{Dehaene:2017} characterize consciousness further. They distinguish unconscious processing (C0) and two orthogonal dimensions of conscious computations: \textit{global availability of information (C1)} and \textit{meta-cognition (C2)}. 

C1 -- global availability -- is a consequence of the distributed organization of the brain as a deep hierarchy of specialized subsystems that must be synchronized and of the need to act, which means that we cannot stick to a diversity of probabilistic interpretations and action options, but must decide in favor of a single course of action. Such decision-making requires efficient pooling over all available sources of information, considering the available action options and selecting the best one, sticking to this choice over time, and coordinating internal and external processes towards the achievement of that subgoal. Attention -- selective processing of information -- is crucial for items entering consciousness, but attention is not limited to conscious processing. Unconscious C0 processing also includes bottom-up and top-down attentional mechanisms, which operate in parallel to prioritize and flexibly route information -- often without bringing it to consciousness. The hierarchical system of sieves that operate unconsciously computes probability distributions, but only a single sample drawn from these becomes conscious at a given time, making it available globally to all specialized modules. Alternative interpretations might become conscious at other points in time, thus, C1 consciousness is causally responsible for our serial information-processing bottleneck. Attention also implements variable binding~\cite{Greff:2020}. The association of information elements to roles in relations and rules is crucial for applying these templates to varying input and, hence, for multi-step inference and systematic generalization.

Consciousness in the second sense (C2) is characterized by the ability to reflexively represent oneself. When making decisions, we feel more or less confident about our choices. Our brain does not only make perception and action decisions, but also estimates its degree of confidence. State estimation and learning also rely on confidence, for example, we weigh existing knowledge versus new evidence, like a Kalman filter~\cite{Tenenbaum:2011}. Error detection is another example of self-monitoring: just after responding, we sometimes realize that we made an error and change our minds. This might be explained by further evidence that arrived after the decision or by slower C2-processing monitoring fast C0 sensory-motor execution. We don’t just have knowledge, but we also know what we don’t know. Such meta-knowledge is crucial for assessing our limits and for learning. 

C1 and C2 are largely orthogonal and complementary dimensions of consciousness. Their joint possession may have synergistic benefits to organisms and robots. Bringing probabilistic metacognitive C2 information into the global C1 workspace allows it to be held over time, integrated into explicit long-term reflection, and shared with others. On the other hand, the possession of an explicit repertoire of one’s own abilities (C2) improves the efficiency with which C1 information is processed. 

\section{The Need for Conscious Robots}

Despite tremendous progress in C0-like deep neural networks trained end-to-end in tasks such as object recognition, video games, and board games, truly human-like learning and thinking machines will need to go beyond current engineering trends in both what they learn and how they learn it~\cite{Lake:BBS2017}. They need to build causal models of their environment that support explanation and understanding and must harness compositionality and learning-to-learn to rapidly acquire and generalize knowledge to new tasks and situations~\cite{ScholkopfLBKKGB:2021}. For this, equipping machines with C1 and C2 conscious processing will be crucial. Upon success, they would behave as if they were conscious; e.g., they would know that they are seeing something, express confidence in it, report it to others, and may even experience the same perceptual illusions as humans.

Of course, \textit{traditional symbolic AI} systems (GOFAI), like Hierarchical Task Network Planners~\cite{NauAIKMWY:2003} and CRAM~\cite{Beetz:2023}, exhibit some of the properties that are associated with conscious System~2 processing, like compositionality. However, such symbol manipulation systems often lack semantic grounding of the higher-level concepts in terms of the lower-level observations and actions. Whereas pure symbolic representations put every symbol at the same distance from every other symbol, learned embeddings represent concepts through a vector of attributes -- with related concepts being close-by and interpolations being meaningful. GOFAI systems often are too rigid to account for real-world data with outliers, etc. Further, GOFAI search and inference are generally intractable and need to be approximated. Here, learning representations together with inference procedures is needed to generate fast habitual C0 behavior. Finally, GOFAI approaches often do not handle uncertainty, which is crucial for partially observable, stochastic environments.

In recent years, \textit{neuro-symbolic} approaches~\cite{GarcezL:2023, Hitzler:2023, MarraDMR:2024} have been proposed that integrate symbolic and subsymbolic representations, inference, and learning. However, hybrid neuro-symbolic systems~\cite{ManhaeveDKDR:2021, NyeTTL21, LambGGPAV20, GlanoisJFWZ0LH22, Zhou:2024, ShindoPDK23, CornelioD24, DeSmet25, TrinhWLHL24}  inherently use different representations and tools for neural and symbolic computations, which are difficult to integrate tightly. 

\textit{Neurocompositional computing}~\cite{SmolenskyMFGG22} is based on the principles of compositionality and continuity. It encodes structures in vectors that are processed by neural networks and shows promising results by quickly learning tasks from small data sets that require systematic generalization. The Differentiable Tree Machine~\cite{SoulosHMCFSG23} compiles high-level symbolic tree operations into subsymbolic matrix operations on tensors. Here, an agent learns to sequentially select tree operations to execute tree transformations with the help of a tree memory. 

Recently, autoregressively trained embodied multimodal models have been used for generating robotic skills such as grasping and placing objects~\cite{BrohanBCCDFGHHH23, Kim24} and for higher levels of robot control~\cite{DriessXSLCIWTVY23}. These models lack System~2 conscious processing, though. They need much data~\cite{ONeillRMGPLPGMJ24} and computing power; and the addressed scenarios are still relatively simple. 

My hypothesis is that using insights from human cognition for the cognitive architectures of robots by incorporating C1 global availability and C2 metacognition will enable the next level of robot capabilities.  

\textit{\bf Because it extends highly successful C0 processing without a change in tools, I am convinced that a bottom-up way towards consciousness-inspired higher-level cognitive functions for service robots is the way to go.} 

\section{Objectives}
\label{sec:Objectives}

My overall goal is to develop methods for learning higher-level cognitive functions for service robots, which go beyond unconscious routine tasks by incorporating conscious processing to cope with novel situations and self-monitor. 

\runinhead{Unconscious perception and control} The System~1\,/\,C0 routine processing directly interacts with the environment and is hence the basis for any higher-level cognitive functions. 
\begin{wrapfigure}{r}{0.5\textwidth}
    \centering\vspace*{-5ex}
         \includegraphics[width=5.5cm]{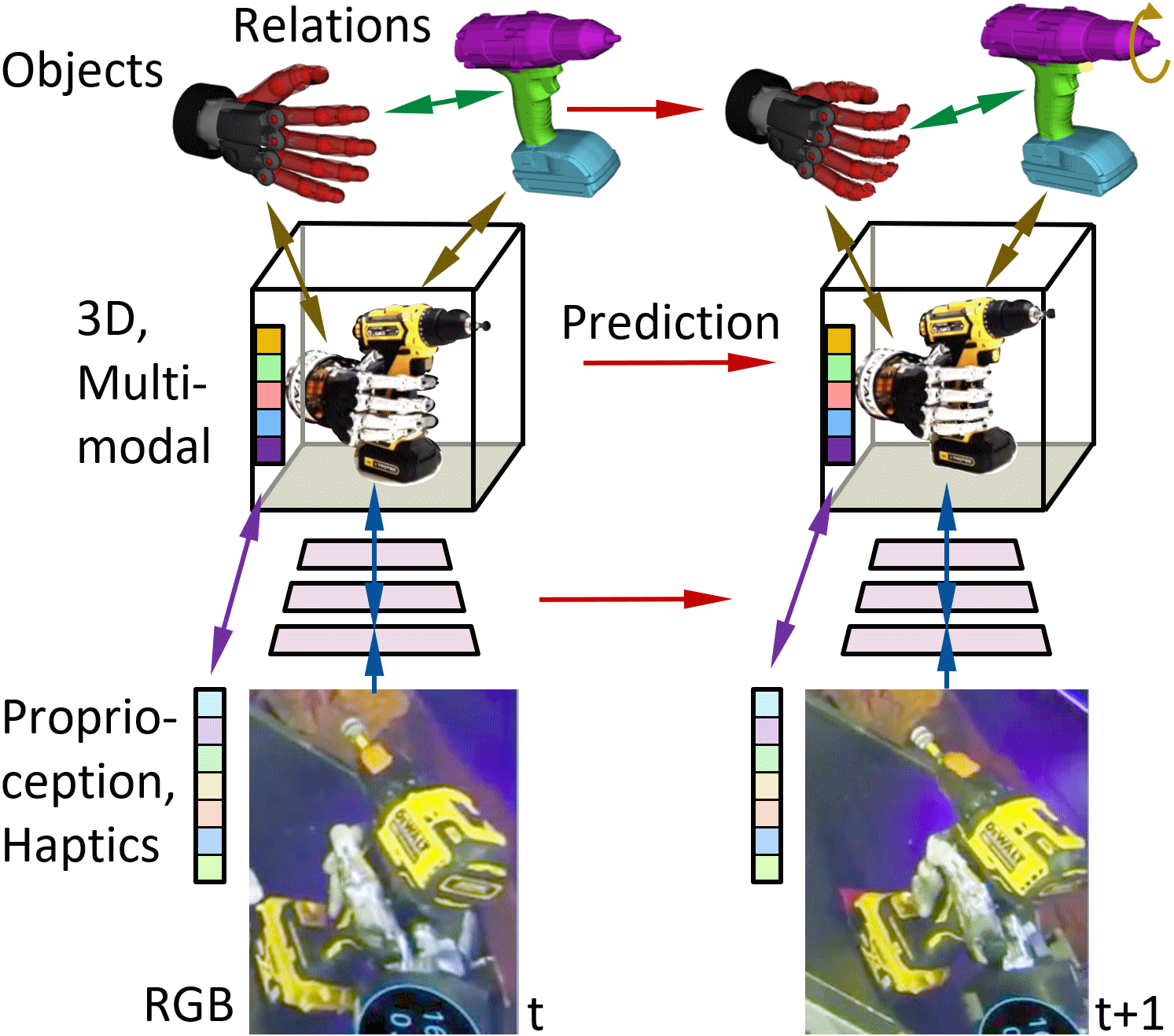}\vspace*{-1ex}
        \caption{Scene perception and prediction on three levels: in the sensor coordinate frame (bottom), in 3D multimodal embeddings (center), and with objects and their relations (top).}
				\vspace*{-2ex}
        \label{fig:PerceptionPrediction}
\end{wrapfigure}
Starting from raw sensory measurements, such as video, depth, forces, and haptics, structured representations of mobile manipulation robot workspaces shall be learned on multiple levels of spatio-temporal abstraction. Abstraction will be realized by coarser spatio-temporal scales and more expressive, sparser representations on the higher levels. The elements of these representations will correspond to increasingly larger entities (parts, objects, groups of objects) in the scene and will be increasingly semantic. The learned representations shall transition from sensor coordinate systems (e.g. the camera frame) to 3D representations of the scene and joint multimodal embeddings. 

They shall model individual objects and the robot end-effectors in their own canonical frame. This will enable learning of category-level shape and appearance spaces within a hierarchical categorization. Scene parsing shall instantiate these models and estimate object parameters, such as pose, shape descriptors, and appearance descriptors. Predictive models for these scene representations shall be learned on all levels of abstraction (see Fig.~\ref{fig:PerceptionPrediction}). Prediction of low-level detail shall be done only for short time horizons. Higher-layer representations shall be predicted with coarser temporal granularity over longer time horizons. These predictions shall be based on individual object dynamics models and on pairwise relational models to account for object interactions, such as contact. The graph of object relations shall be sparsely instantiated according to the relevant object interactions. The predicted representations shall be compared to the feed-forward interpretation of new measurements, such that prediction error can be used to update the representations on all levels. 

The learned predictive models shall be extended by conditioning them on robot actions. This will allow for the rollout of possible futures of robot-environment interaction. Coarse-to-fine model-predictive control of routine skills that do not require conscious attention shall be learned from imagined rollouts on the multiple levels of abstraction. Higher layers shall plan abstract actions longer into the future, which are concretized on lower levels for shorter time horizons. A large variety of skills shall be learned for modular behaviors that activate coarse-to-fine actors according to the situation. Binding objects or places to roles shall yield parametrizable skills, such as grasping or placing an object or navigating towards a waypoint while avoiding obstacles.   

\runinhead{Conscious prediction and planning} Methods for selecting a small set of elements from the highest-level C0 representations and for maintaining them in a working memory (WM) shall be learned. This WM will be the basis for learning action-conditioned predictions, based on binding selected elements to variables of applicable rules. 

\begin{wrapfigure}{r}{0.5\textwidth}
    \centering\vspace*{-4ex}
         \includegraphics[width=5.5cm]{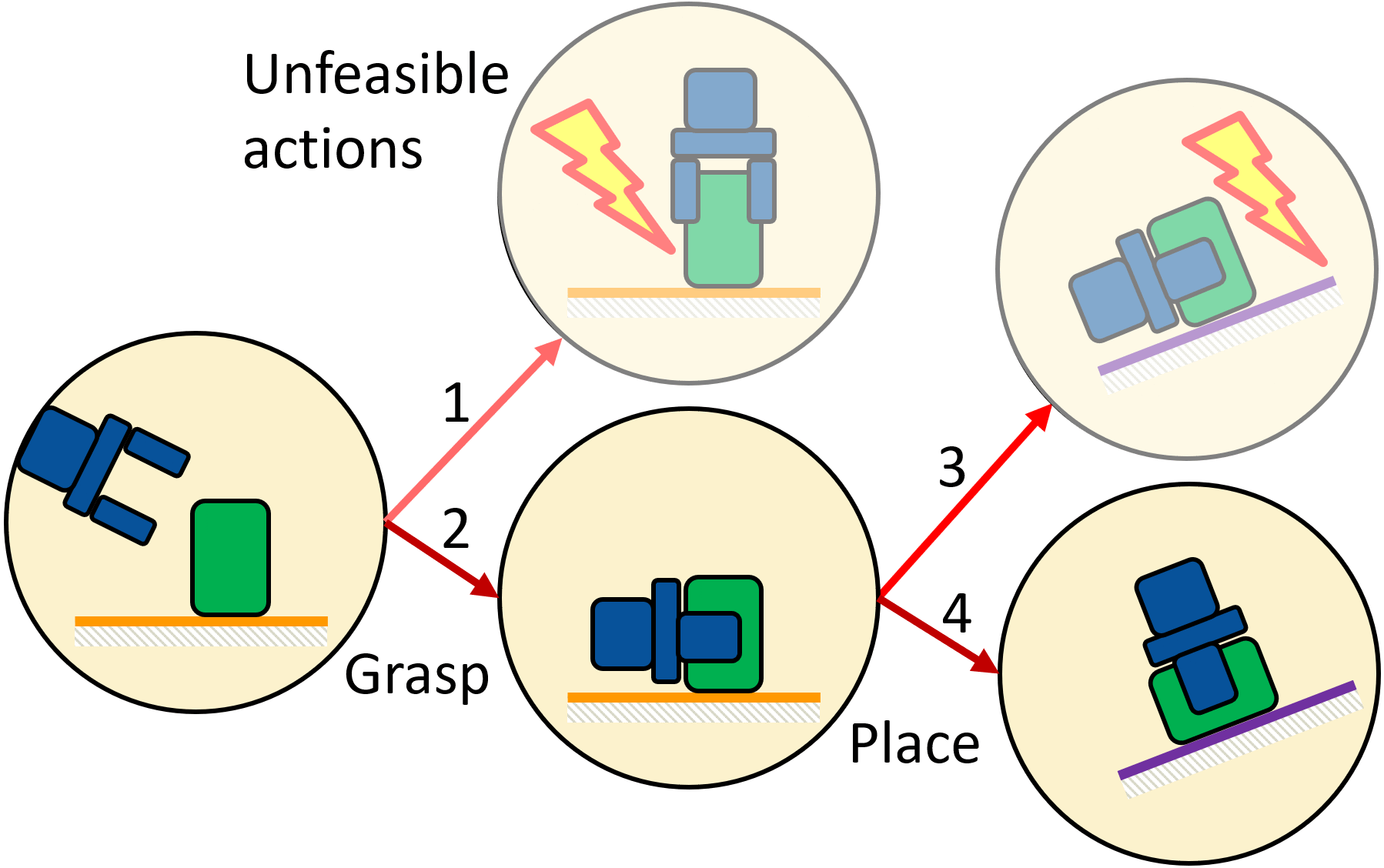}\vspace*{-1ex}
        \caption{Conscious planning. The WM state is rolled out using actions 1--4. Actions 1 and 3 are unfeasible (unreachable top grasp and unstable placement, respectively).}
				\vspace*{-2ex}
        \label{fig:WMPlanning}
\end{wrapfigure}
Structured predictions of state transition rewards, value, and action selection probabilities shall be learned from interactions with simulated and real environments. LLMs shall be incorporated as oracles.

As illustrated in Fig.~\ref{fig:WMPlanning}, the learned WM world models shall be used for efficient action planning by sequential search. A spatio-temporal action abstraction shall be learned, such that sub-plans are reused in different contexts. The learned models shall be used for autonomous operation of mobile manipulation in complex novel situations.

\begin{wrapfigure}{r}{0.35\textwidth}
    \centering\vspace*{-0ex}
         \includegraphics[width=4cm]{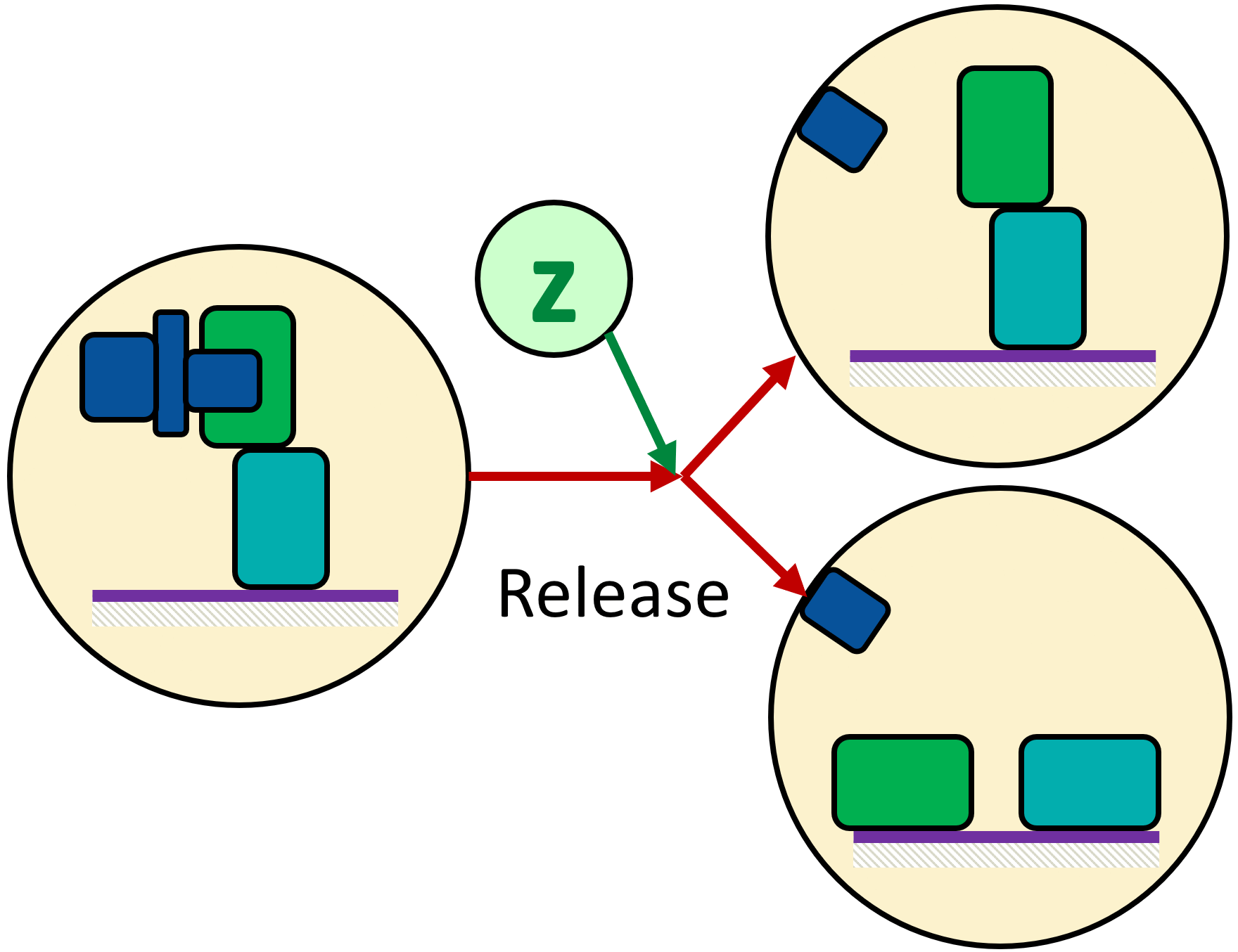}\vspace*{-1ex}
        \caption{Predicting multiple plausible futures conditioned on latent variable $z$.}
				\vspace*{-2ex}
        \label{fig:MultipleFutures}
\end{wrapfigure}

\runinhead{Conscious self-monitoring} Methods for assessing the confidence of perceptions and predictions shall be developed. They shall be based on learning the distributions of latent variables on which multiple plausible futures can be conditioned (see Fig.~\ref{fig:MultipleFutures}). 
By sampling from these variables, a tree-manifold of state-action rollouts can be generated, such that not only average value, but also its variance and worst-case return can be estimated. These quantities shall be incorporated into perception and action selection, to obtain policies that collect more information when needed and avoid dangers. Furthermore, the execution of low-level skills shall be monitored by comparing the current percept to expected outcomes to detect errors and to mitigate them.

\section{Methodology}
\label{sec:Methodology}

My approach will be to add suitable \textit{inductive biases} to deep reinforcement learning (DRL), such that structured representations and conscious processing are enforced, which will enable systematic generalization and self-monitoring. Inductive biases reflect assumptions about the statistics of modeled scenes and robot-environment interactions and are necessary for generalization~\cite{Mitchell80}. For instance, hierarchical convolutional neural networks (CNNs)~\cite{LeCun15} hardwire local dependencies, translation equivariance, hierarchical structure, and invariance to local deformations; whereas recurrent neural networks~\cite{Hochreiter97} exploit equivariance over time. 

\begin{wrapfigure}{r}{0.6\textwidth}
    \centering\vspace*{-0ex}
         \includegraphics[width=6cm]{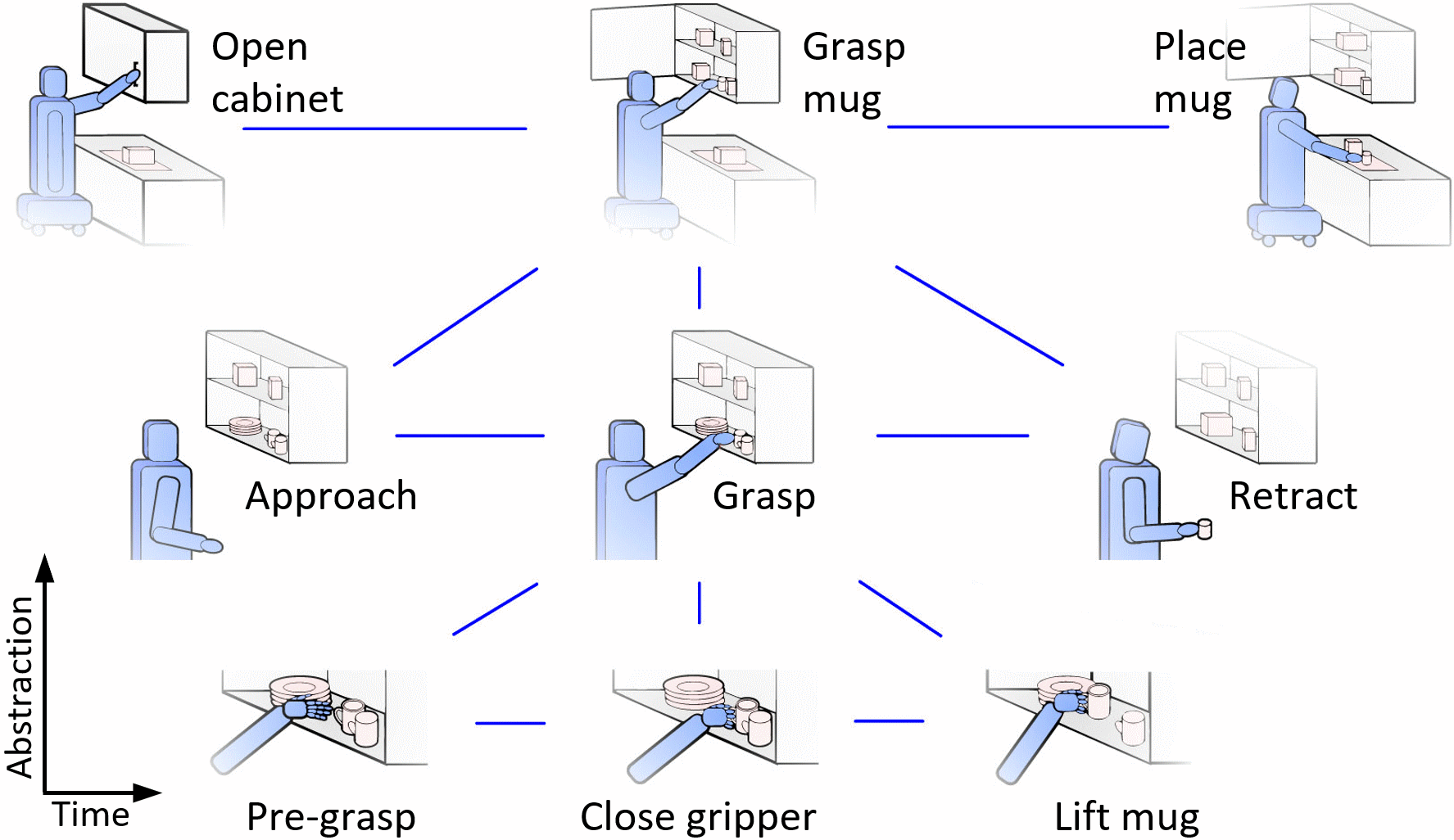}
        \caption{Modeling actions on multiple levels of abstraction.}
				\vspace*{-2ex}
        \label{fig:MultiLevelAction}
\end{wrapfigure}

Further biases are needed for higher cognitive functions. One example of these is choosing the \textit{appropriate frame of reference} for modeling. Commonly, deep neural networks represent visual scenes in a sensor coordinate system. Describing objects in object-centered canonical frames normalizes away the variability induced by the 6D object pose~\cite{Wang0HVSG:CVPR19}. In such object-centered frames, shape and appearance spaces can be learned much easier. Of course, such canonical frames are also useful for individual parts of objects, for which the 6D pose relative to the object-centered frame must be modeled. The \textit{projection of the 3D world to 2D images} induces occlusions and discontinuities at object boundaries. Modeling the scene in 2.5D by individual depth layers or directly in 3D allows for more complete, continuous representations where occluded parts are present and hence unoccluded parts can be predicted. Much of the image motion can be explained by \textit{camera motion}. Hence, approaches that explicitly model the projection from 3D to the variable camera view can represent this dependency compactly. One particularly powerful assumption is \textit{relational inductive bias}~\cite{Battaglia18}. It expresses the observation that scenes can often be described in terms of entities (objects, parts, groups) and their sparse pairwise interactions (relations). Relations describe interactions between entities on adjacent levels of abstraction, e.g. between the whole object and its parts, which represents the \textit{compositional structure} of the world. Such compositional hierarchy can also be found on the action side, where tasks are composed of multiple subtasks and subtasks are composed of individual skills (see Fig.~\ref{fig:MultiLevelAction}). 

Relations are also present between closely interacting entities on the same level of description, e.g. for objects that are in contact or for adjacent subtasks. In \textit{hierarchical categorization}, objects or actions are categorized on multiple granularities~\cite{DengDSLL009}. This allows for pooling instances of multiple finer categories to learn models of coarser categories. \textit{Planning in the now}~\cite{KaelblingL11} refers to the assumption that typical planning problems are not like mazes but can be solved by plans that consist of only a few steps which are described on a detailed, concrete level for the immediate future and on coarser, more abstract levels for the more distant future. This makes planning exponentially more efficient than detailed long-horizon planning.

Crucial is the \textit{consciousness prior} proposed by Bengio~\cite{Bengio17}. It assumes that from subconscious, massively parallel computed representations, a small subset of elements is selected by attention mechanisms for sequential processing. This corresponds to a sparse factor graph on a conceptual, symbolic level, which affords abstract reasoning. Generic factors are probabilistic analogs of logical rules with quantifiers, i.e., with variables or arguments that can be bound~\cite{Goyal22}. On this level, graph neural networks~\cite{WuPCLZY21} are applicable, which exploit \textit{equivariance over entities and relations}. 

\textit{Self-attention} used in transformer networks~\cite{VaswaniSPUJGKP17} provides flexible information routing and learns sparse features with the sample complexity scaling only logarithmically with the context size~\cite{EdelmanGKZ22}. High-level representations that describe verbalizable concepts as semantic variables that play a causal role can be encouraged with an \textit{inductive bias towards words}~\cite{Goyal22}. When planning is restricted to maintaining a single state consisting of few elements and search is serial and can consist of few action-conditioned WM state predictions only, there is \textit{pressure to aggregate} elements to higher-level entities by discovering new concepts and macro actions. 

Inductive biases alone will not suffice to address complex, open-ended real-world domains, because real-robot experience is expensive to obtain and cannot be collected in large enough quantities. Fortunately, \textit{foundation models} for vision~\cite{Dehghani23, Ravi24}, language~\cite{OpenAI23, Anil23}, and multimodal data~\cite{GirdharELSAJM23, HanGZ0ZL24} are available. They have been trained on Internet-scale data and summarize much more experience than real robots could make. Incorporating this knowledge through distillation will be a crucial factor for success. 
Another source of knowledge that I will incorporate is \textit{human guidance and demonstrations}. Still, it will be necessary to develop methods first in a \textit{photorealistic physics-based simulation}~\cite{MittalYYLRHYSGMMBSHG23}, where experience can be made cheaply in large quantities without endangering the real robot, before transferring them to the real world.

\begin{figure}[t]
    \centering \footnotesize \vspace*{-1ex}
         a)\includegraphics[height=4.1cm]{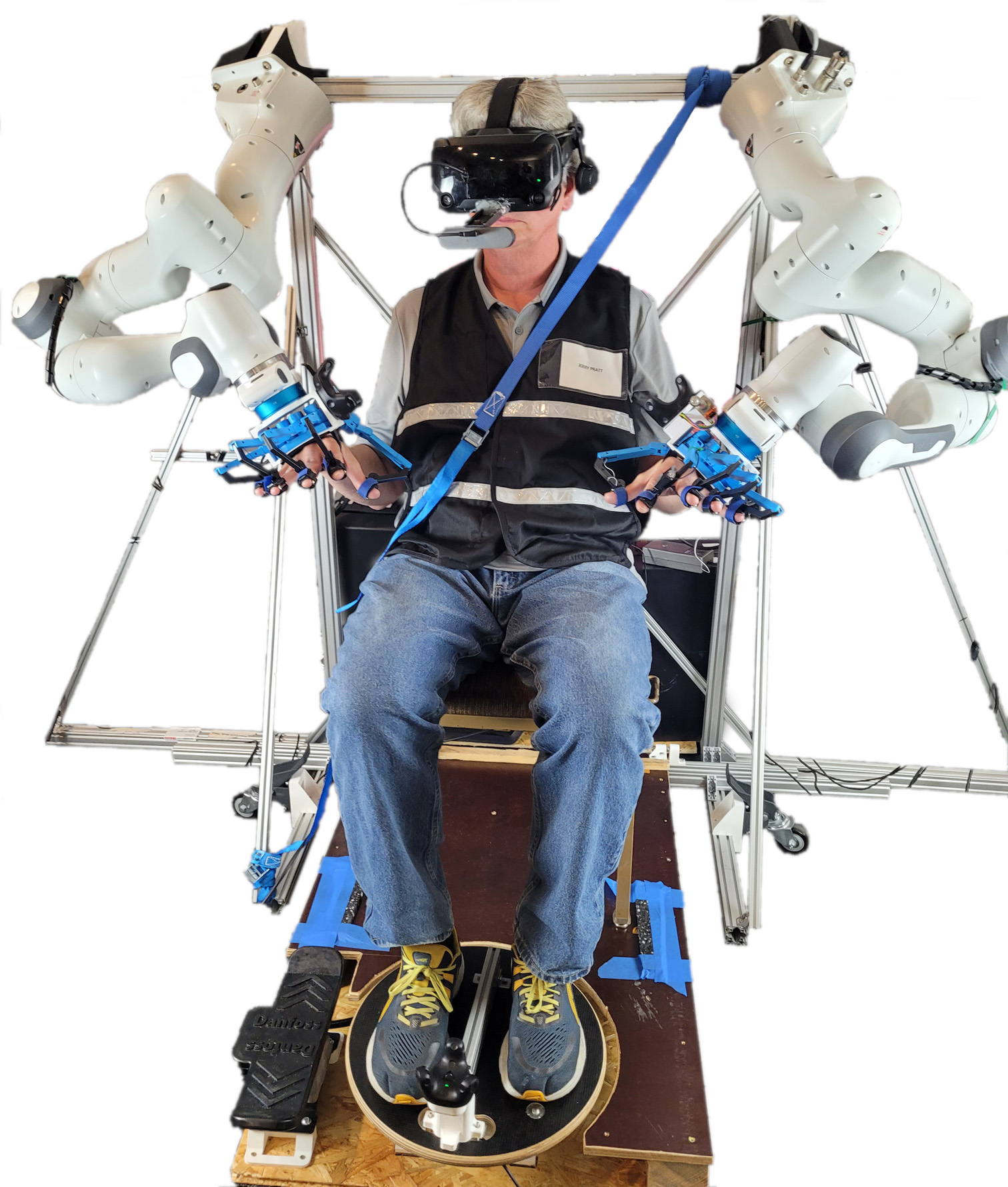}~\includegraphics[height=4.7 cm]{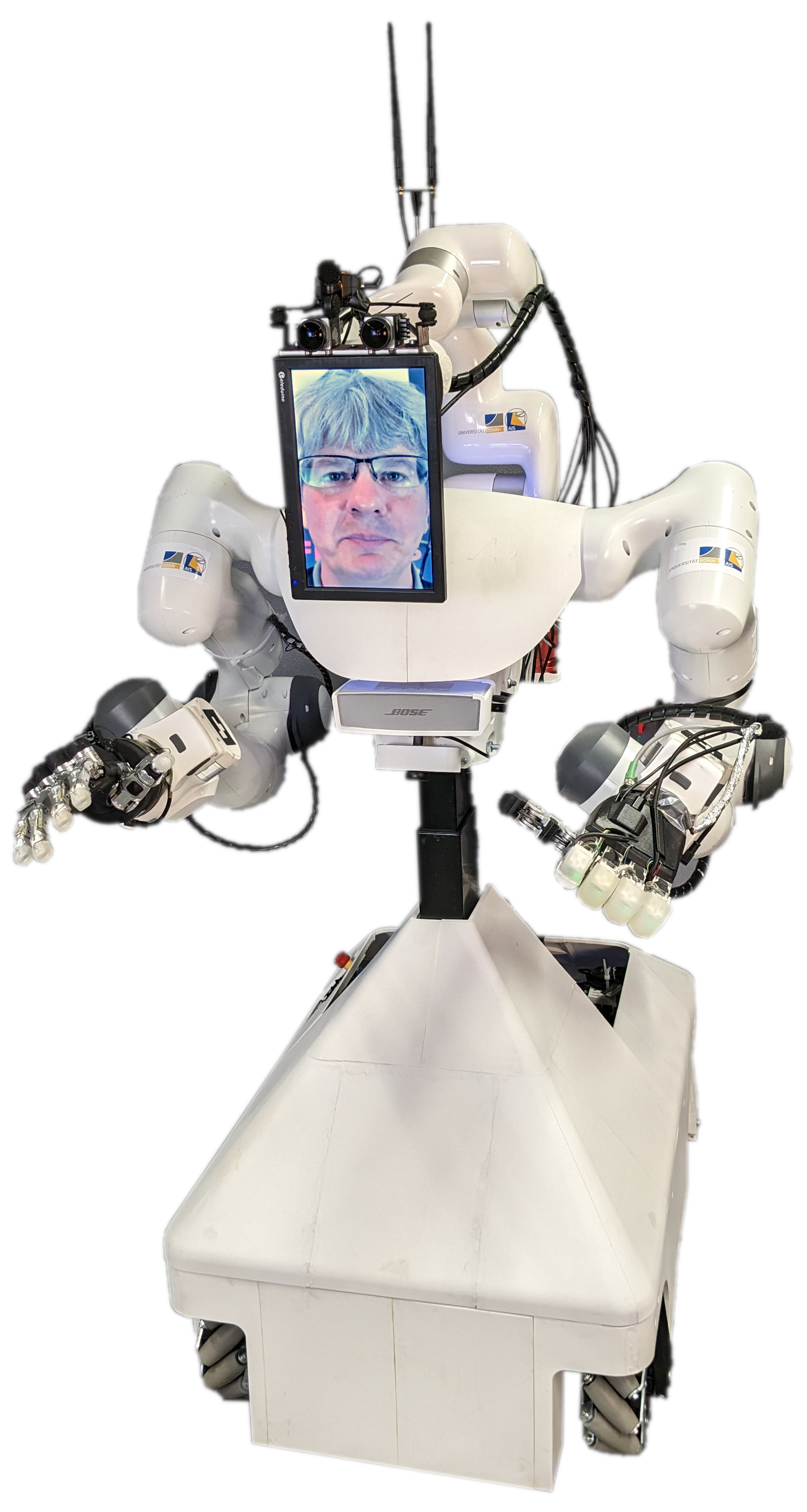}~~~~~~
				 b)\,\includegraphics[height=4.2cm]{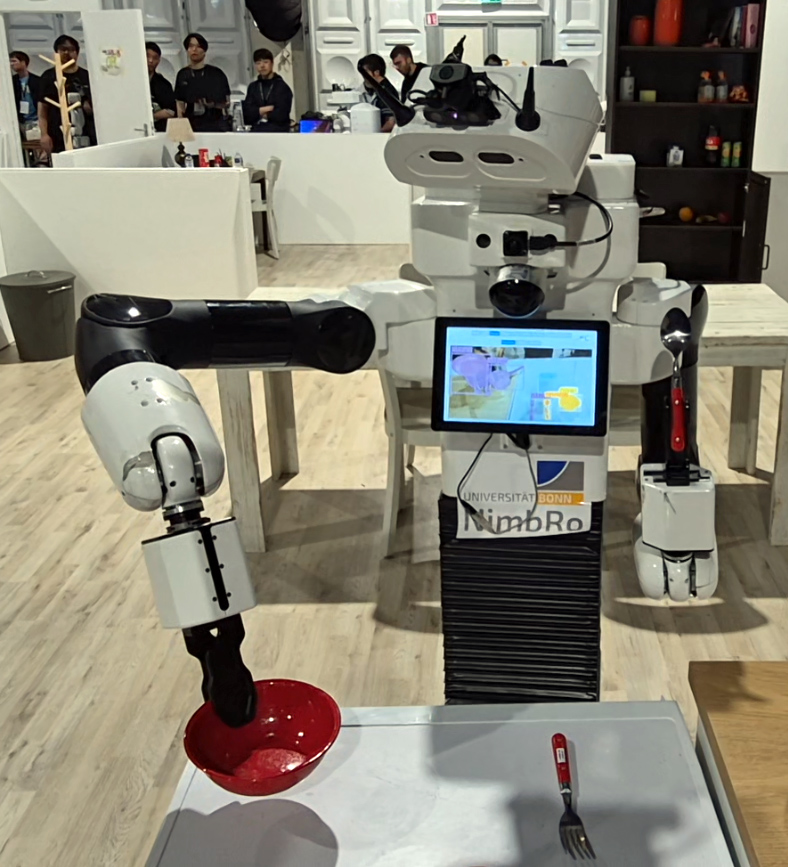}\vspace*{-.1ex}
        \caption{a) NimbRo Avatar system~\cite{LenzSchwarz:SORO2023}; b) NimbRo@Home robot~\cite{Memmesheimer25}.}
        \label{fig:NimbRo}
\end{figure}

I will not approach human-like mobile manipulation in its full generality in a single step, but will increase the level of difficulty gradually along the following dimensions: \textit{Number of objects:} Starting without an object (learning a self-model), proceeding with a single object (grasping, placing, pushing), and finally considering the manipulation of two objects and the use of tools.  \textit{DoF of the robot:} From single-handed object manipulation over bimanual tasks to mobile manipulation tasks.  \textit{DoF of objects:} From rigid objects, to articulated objects, to deformable objects.  \textit{Feedback modalities:} Starting with RGB-D camera-based feedback, adding simple robot state and force-torque sensing, to rich multimodal feedback incorporating haptic measurements and 3D LiDAR.  \textit{Familiarity of objects and tasks:} From known objects and tasks, over variations of known requiring parameter adaptations, to unfamiliar objects and tasks requiring compositional generalization. \textit{Level of abstraction:} Starting with motion control on a fast time scale, proceeding with movement primitives, continuing with skills, such as grasping or placing an object, and finally considering chaining of skills to solve entire tasks.  \textit{Dynamics:} Starting with quasi-static motion by considering only kinematics, proceeding with slow, compliant motion with interaction forces, and finally modeling dynamic effects of fast movements.

Intuitive immersive telepresence systems enable transporting human presence to remote locations in real time. My team NimbRo developed the winning entry for the ANA Avatar XPRIZE competition~\cite{LenzSchwarz:SORO2023} (see Fig.~\ref{fig:NimbRo}a). Telepresence also provides a rich source of environment interaction data for learning structured perception and autonomous behavior. 

My team NimbRo develops perception, planning, and learning for anthropomorphic mobile manipulation robots providing personal assistance~\cite{StucklerSB16} and benchmarks them in the RoboCup@Home league, where we recently won the German Open 2024 and RoboCup 2024 OPL competitions~\cite{Memmesheimer25} (see Fig.~\ref{fig:NimbRo}b).

\section{Conclusions}

By incorporating insights from human cognition, the next generation of service robots will systematically generalize their knowledge to cope with novelty. This new generation of robots will also monitor themselves to obtain more information when needed, to avoid risks, and to detect and mitigate errors. Conscious service robots have much potential for numerous open-ended application domains, including assistance in everyday environments. 
Moreover, artificial conscious processing will contribute to a better understanding of consciousness in humans and other animals.

\bibliographystyle{spmpsci}
\bibliography{references}

\end{document}